\documentclass[letterpaper, 10 pt, conference]{ieeeconf}  

\IEEEoverridecommandlockouts                              

\usepackage{graphics} 
\usepackage{epsfig} 
\usepackage{multirow}
\usepackage{longtable}
\usepackage{booktabs}
\usepackage[sorting=none,citestyle=numeric-comp]{biblatex}
\title{\LARGE \bf
Generating consistent PDDL domains with Large Language Models
}

\author{Pavel Smirnov$^{1}$, Frank Joublin$^{1}$, Antonello Ceravola$^{1}$ and Michael Gienger$^{1}$%
\thanks{$^{1}$Honda Research Institute Europe, Offenbach, Germany 
\newline
        {\tt\small \{firstname.lastname\}@honda-ri.de}}
        }%
        
\begin{document}

\maketitle
\thispagestyle{empty}
\pagestyle{empty}

\begin{abstract}
Large Language Models (LLMs) are capable of transforming natural language domain descriptions into plausibly looking PDDL markup. However, ensuring that actions are consistent within domains still remains a challenging task. In this paper we present a novel concept to significantly improve the quality of LLM-generated PDDL models by performing automated consistency checking during the generation process. Although the proposed consistency checking strategies still can't guarantee absolute correctness of generated models, they can serve as valuable source of feedback reducing the amount of correction efforts expected from a human in the loop. We demonstrate the capabilities of our error detection approach on a number of classical and custom planning domains (logistics, gripper, tyreworld, household, pizza). 
\end{abstract}

\section{Introduction}
Large Language Models got reputation as valuable sources of commonsense knowledge for autonomous agents \cite{xie2023openagents, wang2023survey}. Compared to virtual agents, embodied ones utilize commonsense knowledge for high-level task planning together with low-level manipulation planning. In this context LLMs have already shown to be effective in translating natural language users' requests into tasks for executions on a robot \cite{chen2023autotamp,joublin2023copal,xie2023translating, lin2023text2motion}. 
The Planning Domain Definition Language (PDDL) \cite{aeronautiques1998pddl} is a well-known description formalism for modelling actions as well as their preconditions and effects. Representing actions in PDDL and running a domain independent PDDL-planner allows to generate executable plans, which potentially get converted to executable instructions, or serve as a prediction model about the future development of the situation. PDDL can play a role as a bridge between LLM output and execution systems which opens interesting prospectives for goal reasoning \cite{aha2018goal} and probabilistic thought modelling \cite{wong2023word}. 

In this paper we focus on utilizing LLMs for the automated generation of PDDL-domains out of users' requests expressed in natural language. Although the idea of extracting PDDL-domains with LLMs is not novel \cite{guan2024leveraging}, achieving consistency in extracted domains, so that they can be directly used for planning, still remains a challenge \cite{gragera2023exploring}. Another challenge is to detect and explain inconsistencies of failing PDDL domains, so that that they can be repaired based on a plan which is considered to be valid \cite{gragera2023planning,lin2022planning,lin2023towards}. The contribution of our paper bridges the gap between these two challenges, with the target to solve the problem of generating runnable PDDL domains and problems. The analysis of semantic correctness as well as quality of textual descriptions is out of scope of this paper.

The contribution of the paper is twofold: 1) a generation strategy for avoiding syntax errors in a generated markup; 2) a combination of consistency checks and reachability analysis techniques over LLM-generated domains and problems. In particular, we integrate several mechanisms into a LLM-based generation loop. Misused predicates and types of parameters are getting filtered out before the planning, and missing, contradicting or never reached predicates are getting detected and forwarded back to LLM to correct mistakes. As a result, the amount of errors in the generated domains is reduced, leading to significantly improved domain descriptions for the last stage human checks. 

The remainder of the paper is organized as follows:
in section \ref{sec:problem_statement} we describe the role of the domain description languages in robotics, and define the problem statement. In section \ref{sec:method} we propose a PDDL generation strategy, discuss its internal steps and overview different types of flaws covered by the validation mechanisms. In section \ref{sec:experiments} we describe experiments and provide statistics results about amounts of discovered and resolved mistakes for five planning domains. In section \ref{sec:related_work} we discuss related work and highlight differences between their and our contributions.

%
\section{Problem statement}
\label{sec:problem_statement}
We focus on the PDDL generation problem in the context of automated goal reasoning \cite{aha2018goal, niemueller2019goal}, where an agent is supposed to create and maintain its plans about long-term goals. For example, a service robot is assisting a human in preparing dinner and has to track the progress of intermediate steps, since these could be also unexpectedly performed by a human. To provide such a goal tracking functionality, a fine-granular goal/task modelling formalism has to be chosen (PDDL in our case). Goal progress tracking in this case becomes possible as a result of reasoning over states of objects involved into certain actions. Modelling goals in terms of states of involved objects allows an agent to build a set of goal-related expectations (described as a domain) and perform reasoning (described as a problem) every time states are modified (e.g. if a human performs some unexpected steps). Such a system greatly benefits from dynamically generated PDDL-domains build with commonsense of LLMs. Generating PDDL domains used to be a time consuming task performed by human experts only. Today's LLMs are capable of generating plausible-looking PDDL markup for various domains and problems, although both should be considered as affected by hallucination problems\cite{ji2023survey}. Apart from syntax errors in PDDL markup,  domain descriptions require resolution of internal consistencies. Our target is to enhance LLMs with pre- and post-processing tools, so that the executability of generated PDDL domains can be checked immediately and can serve as explanatory feedback source to guide the LLM towards solving the discovered flaws.   


\section{Method}
\label{sec:method}
\begin{figure}
  \centering
  \includegraphics[width=0.45\textwidth]{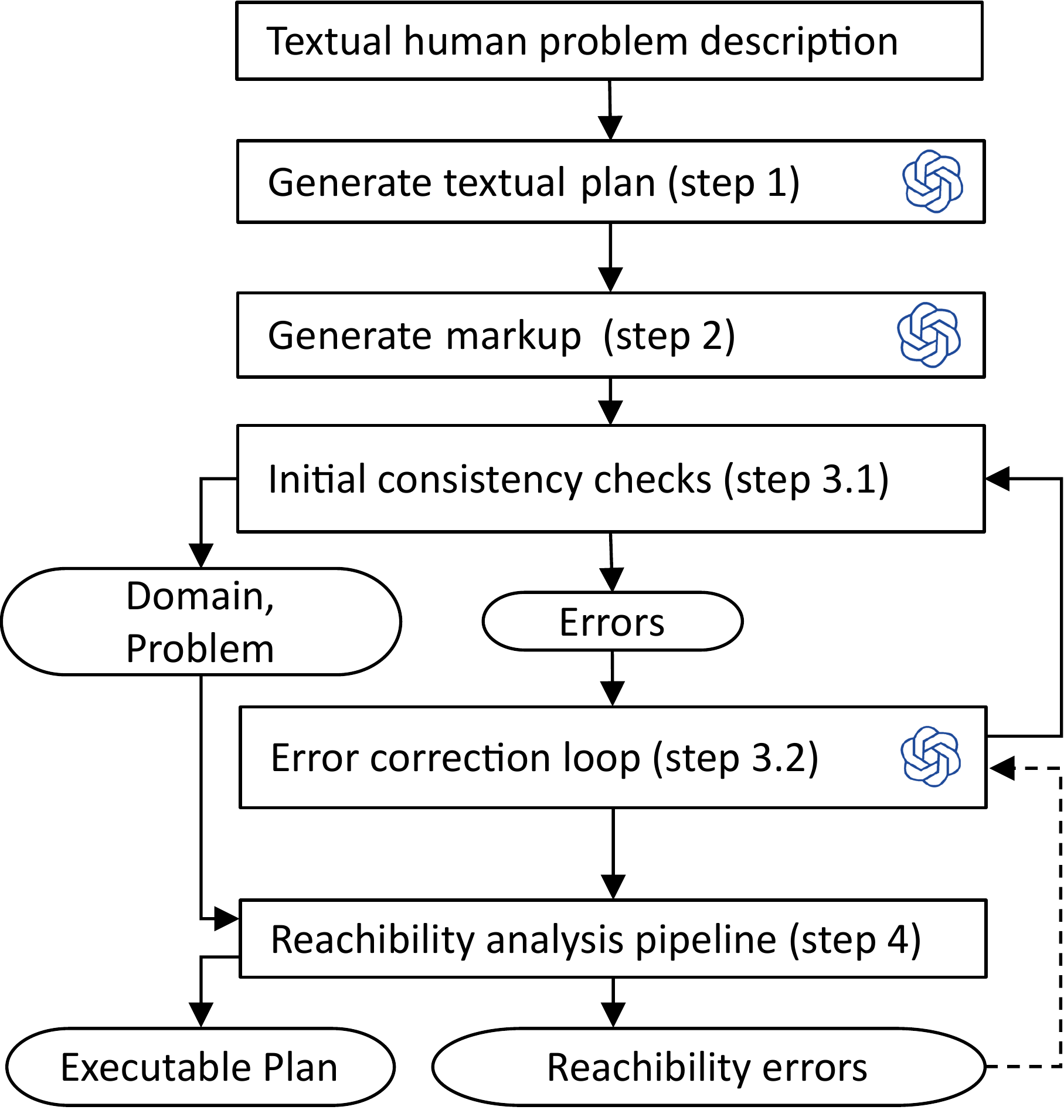}
  \caption{Main generation pipeline}
  \label{fig:main_pipeline}
\end{figure}

Our method was inspired by the idea of generating actions' preconditions and effects\cite{guan2024leveraging}, although it has a number of differences. 
First, we propose a different generation strategy, where domain and problem are prompted as a single model, which gets further corrected. Comparing against per-action generation strategy of the proposed paper, our approach allows to avoid actions aggregation efforts when actions are propmted one-by-one and changes might be required on already generated actions. 
Second, our PDDL domain/problem generation pipeline (see Fig. \ref{fig:main_pipeline}) also includes PDDL-compilation, consistency checking, and goal reachability analysis, which ensures that generated domain/model reaches planning phase and gets either a plan or a feedback about goal's reachability.

Within our generation pipeline we rely on LLM back-prompting techniques \cite{zheng2023step}, which are used for resolving inconsistency errors in dynamically generated domains/problems and increase plan generation success rate. 



\subsection{Generating textual goal plan (step 1)}

The initial user input consists of a shortly-formulated goal (e.g. prepare a pizza), defining sometimes implicitly the initial state of a scene and a target goal-state. The purpose of the first step is to tap into the common sense knowledge of LLM to generate in a chain of thought (CoT) manner
\cite{wei2022chain} a richer textual description of the problem. The LLM gets prompted to use the human initial description, to identify an initial state for the problem, a goal state, and a sequence of actions (to appear in a plan) to reach the goal. It is then instructed to write a list of identified actions and to define for each of them the necessary preconditions and effects. Finally, the LLM is asked to identify the constraints of the problem (e.g. connectivity between locations, actor-action relations, object properties or states, ... ). The textual definition of the domain becomes an input for the next phase. 

\subsection{Generating domain markup (step 2)}
The second phase in the pipeline is the generation of the domain in the form of a target code markup. LLMs are known for being capable of generating code in a requested language (e.g. in form of PDDL-markup). Because of the next-token prediction nature \cite{ji2023survey}, LLMs can't guarantee that generated code is free of syntactic errors. This necessitates the involvement of extra parsing and validation mechanisms. Back-prompting LLMs for fixing those syntactic errors is an unnecessary time-consuming overhead, and might be more beneficial to use it for consistency checking. To reduce syntactic errors, the prompt designer can choose a more widely used format than PDDL (e.g. JSON) since the amount of available examples seen by the LLM during its training is much higher. Similarly, it is wise to use a more commonly used format for the action description, such as function signatures\footnote{https://python-forge.readthedocs.io/en/latest/signature.html}. This will trigger LLMs to use more semantically rich variable names (rather than using PDDL traditional single-letter names) and reduce syntax-specific errors (in case of PDDL those are question marks, dashes and so on). Compilation from function signatures format to a PDDL format is a trivial task, which is performed later in the pipeline. The output of step 2 is a JSON object with a predefined structure similar but syntactically different from PDDL.

\subsection{Initial consistency checks (step 3.1)}
At this step, a syntactically correct JSON markup undergoes the initial consistency checks component which realizes a collection of verifications, as can be seen by the list of errors that it produces (see Table I). This checks consider inconsistency in redundant parts of the markup (e.g. between declaration and usage) as well as types consistencies and to some extent some logical aspects like 'Unusable Initial State Predicate' or 'Unreachable Goal Predicate'.

\subsection{Error correction loop (step 3.2)}
In this step, the errors found in step 3.1 are prompted to a third LLM instance to correct and regenerate the JSON domain markup file. Because of the natural language understanding capabilities of LLMs, the prompt does not only contain the error types and location, like in traditional software design, but also their description and in particular one or more suggestions how to resolve the error. After several iteration loops, the resulting domain markup is then passed to step 4.

\subsection{Reachability analysis (step 4)}
\begin{figure}
  \centering
  \includegraphics[width=0.35\textwidth]{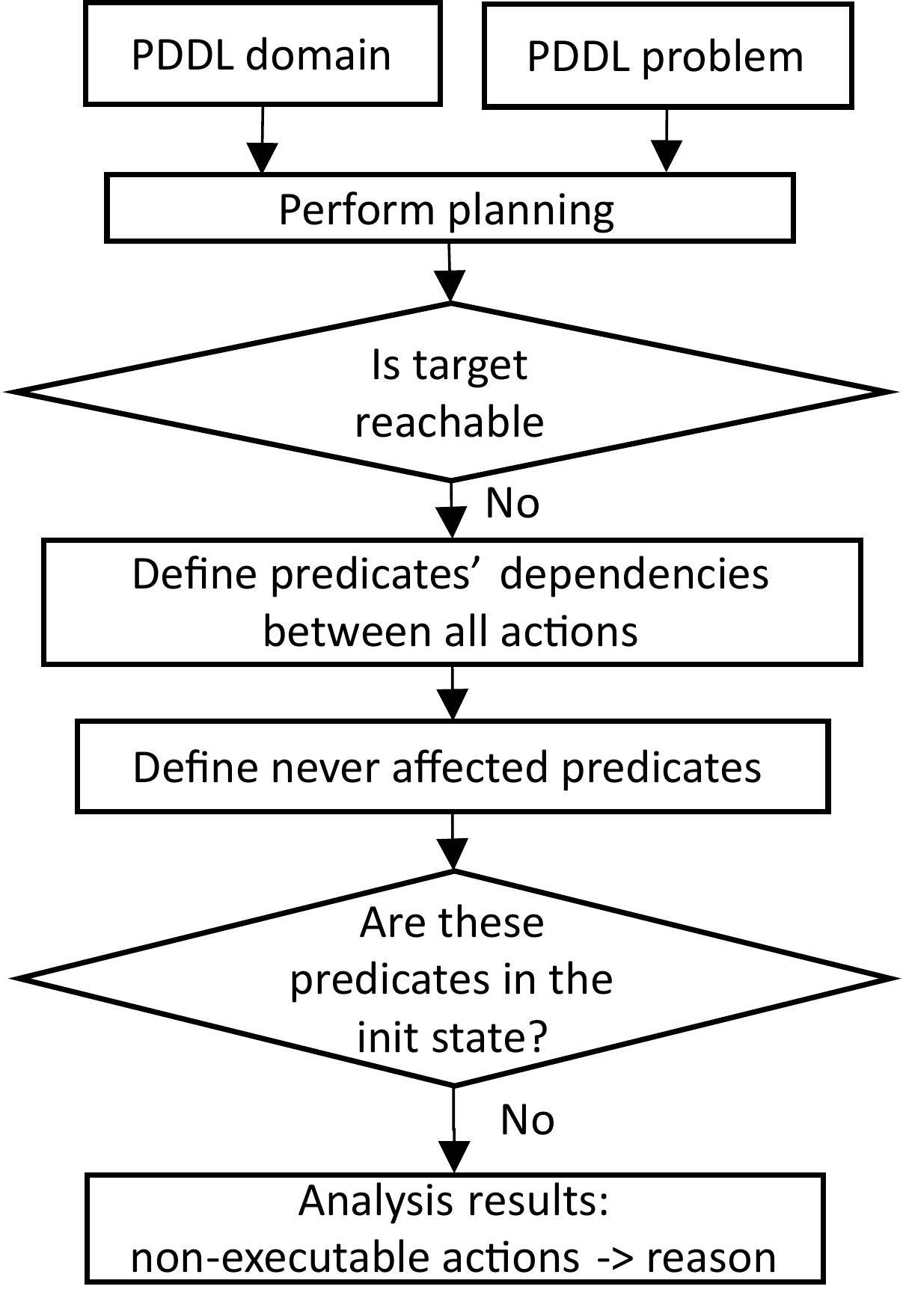}
  \caption{Reachibility analysis pipeline (step 4)}
  \label{fig:reachibility_pipeline} 
\end{figure}

Consistency checks performed at the previous step ensure that outputs (domain and problem) are convertible to a syntactically-valid PDDL markup. However, this does not guarantee that an executable plan will be be generated because of non-reachibility of the goal state. This is addressed in the reachability analysis pipeline (see Fig. \ref{fig:reachibility_pipeline}). The pipeline utilizes a planner (we employed FastDownward \cite{helmert2006fast} with a greedy heuristic), which comes up with a feedback if LLM-generated problem is non-reachable. Non-reachability is usually caused either because of flaws in the domain (e.g. some of the generated actions aren't involved in the plan) or because 
 of missing or incorrectly used predicates in initial or goal state of the problem. Initial information about non-reachable actions or predicates gets extracted out of the planner's translation step \cite{helmert2006fast}. If an action wasn't marked as non-reachable, that does no guarantee that it will be included in a plan due to its preconditions never being satisfiable (e.g. because of the contradicting predicate in preconditions or predicates' types mismatch). For ensuring that, first a dependency analysis between preconditions and effects is performed for all actions. Then, predicates which never get affected by any other action are being searched. At the end, those predicates are searched in the initial state of the problem. In case a predicate is missing or defined with a different type, this gets returned as a non-reachibility feedback for a certain action. For more advanced domain/problem diagnostics techniques we refer the reader to related work section \ref{sec:related_work}. 
It also might happen that the target state is reachable, but not all the generated actions are involved into a plan. In this case non-involved actions are returned as feedback (for the LLM or the human user). In case the plan involves all the generated actions, the domain is considered valid for the specified problem. 
Although, reachability analysis provides valuable insights about  potential planning problem, we didn't make a an iterative correction loop out of this component, but used it in the end of the main pipeline for improving all previous steps.   

\begin{table*}[t]
 \label{table:errors}
  \begin{tabular*}{1.0\textwidth}{ | p{2.5cm} | c | p{5.5cm} | p{7.21cm} | }
\hline
\textbf{Error Type} & \textbf{Rate} & \textbf{Description for LLM} & \textbf{Suggestion for LLM}\\
\hline
Wrong type use & 37.25\% & The type of the parameters used in the predicate 'X' do not match the definition. & Make sure to use parameters of correct type or create a new predicate.\\
\hline
Undeclared parameter use & 26.27\% & The parameter 'X' is not included in the signature of action 'Y'. & Use only parameter of the action or add a new parameter in the action.\\
\hline
Wrong Parameter & 14.62\% & 
(1) In predicate 'Y' the parameter 'X' is a type while a parameter name is expected.\newline\newline
(2) In predicate 'Y' the type 'X' is undefined.\newline
(3) In initial state or goal the parameter 'X' is not defined'.
& 
(1) Either use one of the parameter name defined in the  action signature 'Z' or create a new parameter with type 'X' in the action 'Z' and use it in this predicate 'Y'.\newline
(2) Define a new type 'X' or use an existing one.\newline
(3) Define a new object 'X' or use an existing one.\\
\hline
Missing Parameters & 6.66\% & 
(1) In predicate 'Y' the type 'X' is not defined as parameter name in the action signature 'Z'.\newline
(2) The predicate 'Y' has no parameters.\newline\newline\newline
(3) The action 'Z' has no parameters.
& 
(1) Create the parameter name for the type 'X' and use it in the action signature 'Y'.\newline
(2) Review the need for predicate 'Y' and either remove it if it is not mandatory' or add the parameters from the variable available in the signature of action 'Z'.\newline
(3) Review the need for this action and either remove the action if it is not mandatory', or add the parameters and their types from the variable available in the action's preconditions and effects.\\
\hline
Predicate Mismatch & 5.56\% & The arguments of the predicate 'X' are not matching the types of the predicate definition. & Either remove the mismatched predicate if it is not mandatory or correct the parameter of the predicate 'X' or modify the predicate name to take in account the different type used for this predicate definition since polymorphism is not allowed.\\
\hline
Wrong Type Form & 3.15\% & The type 'X' is not written in upper case.
& Rewrite the type using only upper case letters or underscore.\\
\hline
Missing Predicate & 1.59\% & The predicate 'Y' has not been defined. & Define the predicate 'Y'. \\
\hline
Duplicated Predicate & 1.47\% & The predicate 'Y' has already been defined. & Define a new predicate with a name that take in account the semantic of its argument types. \\
\hline
Unusable Initial State Predicate & 1.47\% & The Initial State predicate 'Y' is not present in the preconditions of any action. & Either remove the predicate from Initial State if is not necessary, or add the predicate in the preconditions of an action that would require this precondition, or create a new action that can have this predicate 'Y' as precondition.\\
\hline
Missing Type & 0.82\% & 
(1) The parameter 'X' do not have a type in the signature of action 'Z'.\newline
(2) The type name 'Y' is not defined.
& 
(1) Add the type of the parameter 'X' in the action signature.\newline\newline
(2) Either create a new type 'Y' or use an existing one.\\
\hline
Object With Multiple Types & 0.61\% & The object 'O' is at least of type 'Y1' and 'Y2'. & Reorganize the type hierarchy of objects to avoid multiple types for an object, or add semantic in the object name to differentiate their belonging types.\\
\hline
Wrong Object Name & 0.25\% & The object 'X' has the same name as type 'X'. & Change the name of the object 'X' to avoid multiple types for an object for example by adding an index to its name.\\
\hline
Unreachable Goal\newline Predicate & 0.16\% & The goal predicate 'Y' is not present in the effect of any action. & Either remove the predicate from the goal states if is not necessary, or add the predicate in the effect of an action that could produce that effect 'Y', or create a new action that can have this predicate 'Y' as an effect.\\
\hline
Duplicated Action & 0.08\% & The action 'Z' has already been defined. & Define a new name for the action 'Z' if it is different, or remove this action if it is redundant.\\
\hline
Duplicated Parameter & 0.04\% & The parameter 'X' is duplicated in signature of action 'Z'. & Change the parameter names to avoid duplicates in the action.\\
\hline
  \end{tabular*}
  \caption{Error types, rates, definitions and suggestions for consistency errors correction loop (step 3 in the main pipeline)}
\end{table*}

\section{Experiments}
\label{sec:experiments}
We use OpenAI GPT-4 (gpt-4-0125-preview) and perform our experiments on a number of well-known classical planning domains (gripper, logistics, tyreworld) as well as custom ones (household \cite{guan2024leveraging} and pizza cooking). For each domain we perform 5 runs and accumulate the following metrics (see table Table II):
\begin{itemize}
    \item Actions count range - amount of actions defined by LLM at step 1 and 2. This is an indication of the problem complexity.
    \item Mean duration of each step in the pipeline (seconds).
    \item Initial error count - mean, minimum and maximum amount of errors in the JSON model markup at the end of step 2 and before correction loop.
    \item Correction iterations count - mean, minimum and maximum amount of correction iterations in step 3.
    \item Rate of non completed generations - percent of runs not finishing with zero errors after 15 iterations.
    \item Rate of non-reachable goals - percentage of generation results, which failed to obtain a plan in step 4, because of reachability issues.
    \item Rate of involved actions - percentage actions involved in a plan (if generated) in step 4. LLM tends to omit important dependencies between actions, so that not all actions are in fact are getting involved into the plan.
\end{itemize}
The scenarios are sorted in the table by increasing complexity. The statistics shows that for simple domains (Gripper, Pizza, Logistics), the framework generates PDDL domains and models, which lead to generated plans. Nevertheless, it does not yet guaranties that semantically correct plan can be generated without human intervention.
More complex domains (Household and Tyreworld) return very non-deterministic results with higher number of flaws. 
The Tyreworld scenario is the hardest one with a range of initial errors count between 90-160 and a rate of non-reachability of 80\%. This indicates that further processing steps and eventually feedback loop would allow to generate better and more executable plans. Nevertheless, sampled manual corrections for the generated domain and problem files are easier and less time consuming (in the range of few minutes) than defining everything from start (in the range of hours).
\label{table:statistics}
\begin{table*}[t]
\begin{tabular*}{1.0\textwidth}{ |p{5.78cm}|c|c|c|c|c| }
\hline
\textbf{Scenario} & \textbf{Gripper} & \textbf{Pizza} & \textbf{Logistics} & \textbf{Household} & \textbf{Tyreworld}  \\ 
 \hline
Actions count range & 3 & 3-7 & 5-6 & 3-7 & 6-12 \\
Mean duration (step 1) [s] & 30.83 $\pm$ 2.68 & 35.97 $\pm$ 6.17 & 33.74 $\pm$ 3.40 & 36.85 $\pm$ 3.43 & 42.16 $\pm$ 8.08\\
Mean duration (step 2) [s] & 37.17 $\pm$ 5.03 & 42.56 $\pm$ 12.81 & 48.12 $\pm$ 5.90 & 39.74 $\pm$ 7.76 & 56.73 $\pm$ 5.69\\
Mean full duration (step 3) [s] & 52.83 $\pm$ 25.49 & 103.30 $\pm$ 71.95 & 155.78 $\pm$ 119.53 & 337.56 $\pm$ 240.62 & 351.97 $\pm$ 243.08\\
Mean duration of correction iteration (step 3) [s] & 20.83 $\pm$ 6.02 & 26.83 $\pm$ 17.32 & 29.77 $\pm$ 15.23 & 59.92 $\pm$ 45.30 & 43.41 $\pm$ 22.69\\
Mean duration (step 4) [s] & 2.04 $\pm$ 0.1 & 1.95 $\pm$ 0.03 & 1.98 $\pm$ 0.02 & 1.74 $\pm$ 0.49 & 1.02 $\pm$ 0.48\\
Initial error count (step 2) & 11.4 $\pm$ 3.67 & 33.2 $\pm$ 32.31 & 40.6 $\pm$ 46.15 & 23.4 $\pm$ 16.32 & 119 $\pm$ 24.15\\
Initial error count range (step 2) & 7-18 & 0-90 & 0-131 & 3-52 & 92-162\\
Correction iterations count (step 3) & 2.4 $\pm$ 0.49 & 3.2 $\pm$ 1.17 & 4.2 $\pm$ 2.48 & 5.4 $\pm$ 3.56 & 9.4 $\pm$ 4.13\\
Correction iterations count range (step 3)& 2-3 & 1-4 & 1-8 & 2-11 & 3-16\\
Rate of non completed generations (step 3) & 0\% & 0\% & 0\% & 0\% & 20\% (1/5)\\
Rate of non-reachable goals (step 4) & 0\% & 0\% & 0\% & 20\% (1/5) & 80\% (4/5)\\
Rate of actions involved in a plan (step 4) & 100\% & 48\% & 50\% & 82\% & 33\%\\
\hline
  \end{tabular*}
  \caption{Experimental evaluation of the proposed pipeline}
\end{table*}

\section{Related work }
\label{sec:related_work}
The idea of utilizing large language models for robotic task planning task is not novel \cite{huang2022language,silver2022pddl,song2023llm}. Backprompting and corrective feedback loops are a helpful way for resolving failing actions preconditions \cite{raman2023cape} or re-planning in case of unexpected failures during execution of the plan \cite{raman2022planning, joublin2023copal}. Compared to LLM-based reasoning over non-structured text inputs, our approach relies on a classical action planning formalism (in this case PDDL), which imposes a higher level of formal control over the agent's actions. However, obtaining a valid PDDL model that exactly describes a given situation is still a challenging task. 

In \cite{wang2024grammar} the authors demonstrate the capabilities of LLMs in generating code in domain specific languages by providing the grammar of a target language in Backus–Naur Form. This is an alternative approach of improving  syntactic validity of LLM's output in a selected markup format. Our approach posits that "function signatures" are sufficient for defining actions and predicates in a basic PDDL-compatible way, unless we need a higher level of expressively, such as loops, conditions, durative-actions, etc. A pythonic way of interacting with LLM is also used in \cite{singh2023progprompt}.

In \cite{guan2024leveraging} the authors utilize LLMs to generate action-related predicates out of natural-language descriptions of a domain. In the input prompt, LLMs get list of actions and their descriptions as well as hierarchy of types to be used. A number of syntactic and type-related checks are performed over LLM outputs per action basis. Authors state that for correcting semantic mistakes, a human-in-the-loop is required. In our paper we are focusing mostly on the domain generation process, where we apply a different prompting strategy - we prompt for the domain together with a problem and perform validations and reachability analysis feedback loop. 



A number of papers addresses the task of identifying flaws in broken PDDL domains. In \cite{lin2023towards} the authors propose an approach for fixing a flawed domain based on a plan which is known to be valid, finding a cardinality - a minimal set for repairs. The authors demonstrate the approach by deleting random predicates. The diagnosis of flawed planning domains is described in \cite{lin2022planning}, where the authors propose a framework for identifying necessary repairs for planning problems. In \cite{gragera2023planning}, the authors propose an approach for fixing compromised actions via  restrictive rules and their costs, utilizing a plan which reaches the goal. Although, our approach relies on structural task analysis techniques, the main focus of our work is in utilizing LLMs as source of commonsense knowledge for generating models for new domains out of textual descriptions - when neither a domain nor a valid problem are defined formally. LLMs tends to make a large variety of consistency errors. The task of detecting them and providing feedback back to LLM is in our focus.

The closest to our work is \cite{gragera2023exploring}, where the authors utilize LLMs for fixing incomplete initial states for domains with incomplete effects in actions. If one of those two situations gets detected, an LLM is asked to provide an alternative solution (full resampling) based on natural language description of the domain and goal-state. The authors conclude that acceptable correction performance demonstrated by ChatGPT could be realized only for the Gripper domain, and for other scenarios are weaken by the LLM introducing errors. Those errors include use of non-existing predicates, omitting type definitions, or omitting relevant tasks. In our approach the algorithm carefully handles all of those error types and utilizes correction loop until the domain and problem become ready to be sent to planning.

In \cite{sakib2024consolidating, sakib2023cooking} the authors propose the generation of reliable execution plans via prompting LLMs multiple times for each problem, and unifying multiple task trees into a graph by discarding incorrect and questionable paths. This method is demonstrated on a cooking robot example, where a recipe was aggregated out of multiple alternative recipes. The accuracy of the method is shown to be higher than the accuracy of GPT-4. Mining plausible goal ingredients across multiple plans is considered for future work, but out of scope of the current paper.

\section{Conclusions}
We propose a novel concept for utilizing commonsense knowledge of LLMs for dynamically generating PDDL domains and problems from natural language descriptions. The approach is based on pre-processing steps, markup generation in a custom format, consistency checks and plan generation. For the cases where domain and problem are syntactically valid, but target goal is not reachable (from the planner point of view) - a detailed feedback pointing to a predicates of a particular action is generated. The same applies to cases of generated plans that do not make use of all actions. Although, we didn't include plan-generation feedback into a correction loop, we've used it during human analysis and improvement of all previous steps of the pipeline. The main contribution of this work is a reduction of the amount of load that is imposed on a human in the loop (compilation and plan generation is done in automated automated). 
We evaluated the capabilities of our error detection approach on a number of classical and custom planning domains (logistics, gripper, tyreworld, household, pizza) and showed that they could be solved to a large extent.

One of the the limitations of this work is the lack of analysis of the semantic correctness of generated plans (e.g. if all required actions have been involved into a plan), which by now is performed by a human in the loop (assisted by results of reachability analysis). For future work we consider to analyse the semantics using LLM-based agents (the direction proposed in \cite{gragera2023exploring}) or mining correct semantics out of multiple instances of generated plans (the direction proposed in \cite{sakib2024consolidating}). 
%



\printbibliography

@article{aha2018goal,
  title={Goal reasoning: Foundations, emerging applications, and prospects},
  author={Aha, David W},
  journal={AI Magazine},
  volume={39},
  number={2},
  pages={3--24},
  year={2018}
}

@article{wong2023word,
  title={From Word Models to World Models: Translating from Natural Language to the Probabilistic Language of Thought},
  author={Wong, Lionel and Grand, Gabriel and Lew, Alexander K and Goodman, Noah D and Mansinghka, Vikash K and Andreas, Jacob and Tenenbaum, Joshua B},
  journal={arXiv preprint arXiv:2306.12672},
  year={2023}
}

@article{aeronautiques1998pddl,
  title={Pddl| the planning domain definition language},
  author={Aeronautiques, Constructions and Howe, Adele and Knoblock, Craig and McDermott, ISI Drew and Ram, Ashwin and Veloso, Manuela and Weld, Daniel and SRI, David Wilkins and Barrett, Anthony and Christianson, Dave and others},
  journal={Technical Report, Tech. Rep.},
  year={1998}
}

@article{guan2024leveraging,
  title={Leveraging pre-trained large language models to construct and utilize world models for model-based task planning},
  author={Guan, Lin and Valmeekam, Karthik and Sreedharan, Sarath and Kambhampati, Subbarao},
  journal={Advances in Neural Information Processing Systems},
  volume={36},
  year={2024}
}

@article{xie2023translating,
  title={Translating natural language to planning goals with large-language models},
  author={Xie, Yaqi and Yu, Chen and Zhu, Tongyao and Bai, Jinbin and Gong, Ze and Soh, Harold},
  journal={arXiv preprint arXiv:2302.05128},
  year={2023}
}

@article{helmert2006fast,
  title={The fast downward planning system},
  author={Helmert, Malte},
  journal={Journal of Artificial Intelligence Research},
  volume={26},
  pages={191--246},
  year={2006}
}

@misc{joublin2023copal,
      title={CoPAL: Corrective Planning of Robot Actions with Large Language Models}, 
      author={Frank Joublin and Antonello Ceravola and Pavel Smirnov and Felix Ocker and Joerg Deigmoeller and Anna Belardinelli and Chao Wang and Stephan Hasler and Daniel Tanneberg and Michael Gienger},
      year={2023},
      eprint={2310.07263},
      archivePrefix={arXiv},
      primaryClass={cs.RO}
}

@inproceedings{silver2022pddl,
  title={PDDL planning with pretrained large language models},
  author={Silver, Tom and Hariprasad, Varun and Shuttleworth, Reece S and Kumar, Nishanth and Lozano-P{\'e}rez, Tom{\'a}s and Kaelbling, Leslie Pack},
  booktitle={NeurIPS 2022 Foundation Models for Decision Making Workshop},
  year={2022}
}

@article{wang2024grammar,
  title={Grammar prompting for domain-specific language generation with large language models},
  author={Wang, Bailin and Wang, Zi and Wang, Xuezhi and Cao, Yuan and A Saurous, Rif and Kim, Yoon},
  journal={Advances in Neural Information Processing Systems},
  volume={36},
  year={2024}
}

@inproceedings{raman2022planning,
  title={Planning with large language models via corrective re-prompting},
  author={Raman, Shreyas Sundara and Cohen, Vanya and Rosen, Eric and Idrees, Ifrah and Paulius, David and Tellex, Stefanie},
  booktitle={NeurIPS 2022 Foundation Models for Decision Making Workshop},
  year={2022}
}

@inproceedings{song2023llm,
  title={Llm-planner: Few-shot grounded planning for embodied agents with large language models},
  author={Song, Chan Hee and Wu, Jiaman and Washington, Clayton and Sadler, Brian M and Chao, Wei-Lun and Su, Yu},
  booktitle={Proceedings of the IEEE/CVF International Conference on Computer Vision},
  pages={2998--3009},
  year={2023}
}

@article{sakib2024consolidating,
  title={Consolidating Trees of Robotic Plans Generated Using Large Language Models to Improve Reliability},
  author={Sakib, Md Sadman and Sun, Yu},
  journal={arXiv preprint arXiv:2401.07868},
  year={2024}
}

@article{sakib2023cooking,
  title={From Cooking Recipes to Robot Task Trees--Improving Planning Correctness and Task Efficiency by Leveraging LLMs with a Knowledge Network},
  author={Sakib, Md Sadman and Sun, Yu},
  journal={arXiv preprint arXiv:2309.09181},
  year={2023}
}

@article{chen2023autotamp,
  title={AutoTAMP: Autoregressive Task and Motion Planning with LLMs as Translators and Checkers},
  author={Chen, Yongchao and Arkin, Jacob and Zhang, Yang and Roy, Nicholas and Fan, Chuchu},
  journal={arXiv preprint arXiv:2306.06531},
  year={2023}
}

@article{ji2023survey,
  title={Survey of hallucination in natural language generation},
  author={Ji, Ziwei and Lee, Nayeon and Frieske, Rita and Yu, Tiezheng and Su, Dan and Xu, Yan and Ishii, Etsuko and Bang, Ye Jin and Madotto, Andrea and Fung, Pascale},
  journal={ACM Computing Surveys},
  volume={55},
  number={12},
  pages={1--38},
  year={2023},
  publisher={ACM New York, NY}
}

@inproceedings{raman2023cape,
  title={Cape: Corrective actions from precondition errors using large language models},
  author={Raman, Shreyas Sundara and Cohen, Vanya and Paulius, David and Idrees, Ifrah and Rosen, Eric and Mooney, Ray and Tellex, Stefanie},
  booktitle={2nd Workshop on Language and Robot Learning: Language as Grounding},
  year={2023}
}

@article{xie2023openagents,
  title={Openagents: An open platform for language agents in the wild},
  author={Xie, Tianbao and Zhou, Fan and Cheng, Zhoujun and Shi, Peng and Weng, Luoxuan and Liu, Yitao and Hua, Toh Jing and Zhao, Junning and Liu, Qian and Liu, Che and others},
  journal={arXiv preprint arXiv:2310.10634},
  year={2023}
}

@inproceedings{singh2023progprompt,
  title={Progprompt: Generating situated robot task plans using large language models},
  author={Singh, Ishika and Blukis, Valts and Mousavian, Arsalan and Goyal, Ankit and Xu, Danfei and Tremblay, Jonathan and Fox, Dieter and Thomason, Jesse and Garg, Animesh},
  booktitle={2023 IEEE International Conference on Robotics and Automation (ICRA)},
  pages={11523--11530},
  year={2023},
  organization={IEEE}
}

@inproceedings{huang2022language,
  title={Language models as zero-shot planners: Extracting actionable knowledge for embodied agents},
  author={Huang, Wenlong and Abbeel, Pieter and Pathak, Deepak and Mordatch, Igor},
  booktitle={International Conference on Machine Learning},
  pages={9118--9147},
  year={2022},
  organization={PMLR}
}

@inproceedings{lin2023towards,
  title={Towards automated modeling assistance: An efficient approach for repairing flawed planning domains},
  author={Lin, Songtuan and Grastien, Alban and Bercher, Pascal},
  booktitle={Proceedings of the 37th AAAI Conference on Artificial Intelligence (AAAI 2023). AAAI Press},
  year={2023}
}

@article{lin2023text2motion,
  title={Text2motion: From natural language instructions to feasible plans},
  author={Lin, Kevin and Agia, Christopher and Migimatsu, Toki and Pavone, Marco and Bohg, Jeannette},
  journal={arXiv preprint arXiv:2303.12153},
  year={2023}
}

@article{wang2023survey,
  title={A survey on large language model based autonomous agents},
  author={Wang, Lei and Ma, Chen and Feng, Xueyang and Zhang, Zeyu and Yang, Hao and Zhang, Jingsen and Chen, Zhiyuan and Tang, Jiakai and Chen, Xu and Lin, Yankai and others},
  journal={arXiv preprint arXiv:2308.11432},
  year={2023}
}

@inproceedings{gragera2023planning,
  title={A planning approach to repair domains with incomplete action effects},
  author={Gragera, Alba and Fuentetaja, Raquel and Garc{\'\i}a-Olaya, {\'A}ngel and Fern{\'a}ndez, Fernando},
  booktitle={Proceedings of the International Conference on Automated Planning and Scheduling},
  volume={33},
  number={1},
  pages={153--161},
  year={2023}
}

@inproceedings{lin2022planning,
  title={Planning Domain Repair as a Diagnosis Problem},
  author={Lin, Songtuan and Grastien, Alban and Bercher, Pascal},
  booktitle={33rd International Workshop on Principle of Diagnosis--DX 2022},
  year={2022}
}

@inproceedings{niemueller2019goal,
  title={Goal reasoning in the clips executive for integrated planning and execution},
  author={Niemueller, Tim and Hofmann, Till and Lakemeyer, Gerhard},
  booktitle={Proceedings of the International Conference on Automated Planning and Scheduling},
  volume={29},
  pages={754--763},
  year={2019}
}

@article{gragera2023exploring,
  title={Exploring the Limitations of using Large Language Models to Fix Planning Tasks},
  author={Gragera, Alba and Pozanco, Alberto},
  year={2023}
}

@article{wei2022chain,
  title={Chain-of-thought prompting elicits reasoning in large language models},
  author={Wei, Jason and Wang, Xuezhi and Schuurmans, Dale and Bosma, Maarten and Xia, Fei and Chi, Ed and Le, Quoc V and Zhou, Denny and others},
  journal={Advances in neural information processing systems},
  volume={35},
  pages={24824--24837},
  year={2022}
}

@inproceedings{zheng2023step,
  title={Step-Back Prompting Enables Reasoning Via Abstraction in Large Language Models},
  author={Zheng, Huaixiu Steven and Mishra, Swaroop and Chen, Xinyun and Cheng, Heng-Tze and Chi, Ed H and Le, Quoc V and Zhou, Denny},
  booktitle={The Twelfth International Conference on Learning Representations},
  year={2023}
}
\end{document}